# Deep Features For Training Support Vector Machines

Loris Nanni [1,*], Stefano Ghidoni[1], Sheryl Brahnam[2]

[1] DEI, University of Padova; loris.nanni@unipd.it and stefano.ghidoni@unipd.it
[2] ITC, Missouri State University; sbrahnam@missouristate.edu

* Correspondence: loris.nanni@unipd.it

**Abstract:** Features play a crucial role in computer vision. Initially designed to detect salient elements by means of handcrafted algorithms, features now are often learned by different layers in Convolutional Neural Networks (CNNs). This paper develops a generic computer vision system based on features extracted from trained CNNs. Multiple learned features are combined into a single structure to work on different image classification tasks. The proposed system was derived by testing several approaches for extracting features from the inner layers of CNNs and using them as inputs to SVMs that are then combined by sum rule. Dimensionality reduction techniques are used to reduce the high dimensionality of inner layers. The resulting generic vision system is shown to significantly boost the performance of standard CNNs across a large and diverse collection of image data sets. An ensemble of different topologies using the same approach obtains state-of-the-art results on an image virus data set.

**Keywords:** Deep learning; transfer learning; non-handcrafted features; texture descriptors; texture classification; ensemble of descriptors.

## 1. Introduction

Extracting salient descriptors from images is the mainstay of many computer vision systems. Typically, these handcrafted descriptors are tailored to overcome specific problems in image classification, the goal being to achieve the best classification accuracy possible while maintaining computational efficiency. Some descriptors, such as Scale Invariant Feature Transform (SIFT) [1], are valued for their robustness, but they can be too computationally expensive for practical purposes. As a consequence, variants of popular handcrafted descriptors, such as some fast variants of SIFT [2], continue to be created in an attempt to overcome inherent shortcomings.

In contrast to computer vision systems that rely on the extraction of handcrafted descriptors are those that depend on deep learners [3], as exemplified in computer vision by Convolutional Neural Networks (CNNs). Deep learning involves designing complex networks composed of specialized layers, and the descriptors or features calculated by these layers are learned from the training samples [4]. Layers in deep learners, like CNN, are known to discover many low-level representations of the data in the early stages that become useful to subsequent layers in charge of providing higher-level features representing the semantics of the data [5]. Close to the input, edges and texture are usually detected [6]. Higher up, features like contours and image patches are discerned. Layer by layer, representations of the data in deep learners become more and more complex. An advantageous characteristic of these deep features is that they are generalizable. Once extracted, they can be treated like other handcrafted features in traditional computer vision systems and applied to many different image problems.

Interest in research investigating feature sets extracted from different layers of pretrained CNNs has grown in recent years. Lower level features extracted from sets of CNN topologies have been explored in [7] and [8] and top layers in [9] and [10]. In [9], for example, images are represented as strings of CNN features with similarities compared with novel distance measures. Convolutional features are extracted in [11], where they are used as a filter bank. In [12], deep activation features are extracted from local patches at multiple scales, with convolutional features taken from the seventh layer of a CNN trained on ImageNet. In [13] and [14], features are extracted from the last convolutional layers of a CNN and in [14] combined with the fully connected (FC) layer. In [15], images are represented using five convolutional layers and two FC layers. Similarly, in [16] convolutional features are extracted from multiple layers combined with FC features. In [17], features are extracted from the penultimate layer of pre-trained CNNs and merged with the outputs of deep layers as well as with CNN scores. Finally, in [18], features are investigated layer by layer and discovered to provide quality information about the texture of images at multiple depths.

This work aims to exploit both the deeper and shallower layers of pre-trained CNNs for representing images with fixed-length feature vectors that can then be trained on a set of Support Vector Machines (SVMs) [19]. Extracting features from the inner layers of CNNs poses a difficulty because they are characterized by high dimensionality, making them unsuitable for training statistical classifiers like SVM. To reduce dimensionality, experiments are run that test the following approaches:

- Classic dimensionality reduction methods: viz., discrete cosine transform (DCT) and principal component analysis (PCA);
- Feature selection approaches (chi-square feature selection);
- Texture descriptors extracted from local binary patterns (LBP) followed by feature selection;
- Co-occurrence among elements of the channels of inner layers;
- Global pooling measurements.

Experiments demonstrate that combining feature sets extracted from inner and outer CNN layers and applying as many different dimensionality reduction techniques as needed obtains close to if not state-of-the-art results on an extensive collection of cross-domain image data sets. In addition, an ensemble of different topologies (DenseNet201 and ResNet50) is tested for virus classification to test generalizability, and this ensemble obtains state-of-the-art results. Performance differences are verified using the Wilcoxon signed-rank test, and all experiments can be replicated using The MATLAB source code available at https://github.com/LorisNanni.

**2. Methods**

*2.1 Feature extraction from convolutional neural networks*

In this work, we extract features from Convolutional Neural Networks [20] pre-trained on the ImageNet dataset [21]. These features are taken from multiple layers of a CNN and then individually trained on separate SVMs (see Figure 1). The CNN architectures investigated in this study are GoogleNet (Inception IV) [22], ResNet50 [23], and Densenet201 [24]. GoogleNet, winner of the ImageNet Large-Scale Visual Recognition Challenge (ILSVRC) in 2014, is a CNN with twenty-two layers. To create deeper layers, GoogleNet uses 1 × 1 convolution and global average pooling. ResNet50, the winner of the ILSVRC2015 contest, is a CNN with fifty layers. To overcome the vanishing gradient problem with deep networks, ResNet incorporates a residual connection. DenseNet201 is extremely deep with 201 layers. This architecture replaces the residual connection with densely connected convolutional layers that are concatenated rather than added to each other as with ResNet. All layers are interconnected in DenseNet, a technique that produces strong gradient flow and that shares low-level information across the entire network.

Unlike many other studies focused on the extraction of features from the output layer, we examine features extracted from deeper layers, as in [17]. The layers considered for extracting features are selected starting from the middle layer of the network and then by considering one layer after every ten, going toward the output layer, with the last four layers always considered. Since deep layers encode high-dimension features, dimensionality reduction methods are also used, as shown in Figure 1, depending on the feature size (i.e., when more than 5000 features are extracted by a given layer).

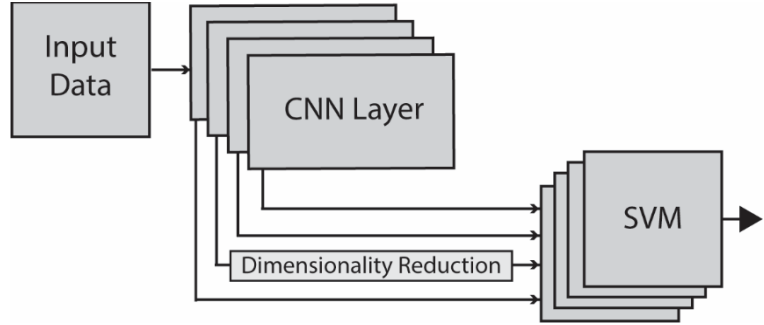

**Figure 1.** Feature extraction from inner layers. The output of each layer is treated as a feature vector, with a dimensionality reduction method applied depending on the vector size (< 5000). All vectors are then processed by a separate SVM and summed for a final decision.

All configurations are investigated considering the possible combinations of the following elements:
- Tuning (with/without): either the CNN used to extract features is pre-trained on ImageNet without any tuning, or it is tuned on the given training set;
- Scope of dimensionality reduction (local/global): either dimensionality reduction is performed separately on each channel of a layer (with results combined), or reduction is applied to the whole layer;
- PCA postprocessing (with/without): either PCA projection is performed after dimensionality reduction, or PCA is not applied.

The dimensionality reduction methods considered in this work are presented in the remainder of this section.

*2.2. Feature Reduction Transforms (PC and DC)*

Dimensionality is reduced by applying two classic transforms: PCA and DCT. In the experimental section, PCA is labeled as PC and DCT as DC.

PCA [25] is a well-known unsupervised technique that projects high-dimensional data into a lower-dimensional subspace. This is accomplished by mapping the original feature vectors into a smaller number of uncorrelated directions that preserve the global Euclidean structure.

The DCT transform [26] balances information packing and computational complexity. DCT components tend to be small in magnitude because the most important information lies in the coefficients with low frequencies. As with PCA, removing small coefficients produces small errors when the transform is reversed to reconstruct the original images.

*2.3. Chi-Square Feature Selection (CHI)*

Univariate feature ranking for classification using chi-square tests is a popular feature selection method. In the experimental section, CHI is the label used for chi-square feature selection.

The chi-square test in statistics tests the independence between two events A and B. If P(AB) = P(A)P(B), then the two events are said to be independent. The same holds when P(A|B) = P(A) and P(B|A) = P(B).

The formula for the chi-square test is
$$X_c^2 = \sum (O_i - E_i)^2 / E_i, \tag{1}$$

where $c$ is the degrees of freedom, $O$ is the observed values, and $E_i$ are the expected values. The degrees of freedom are the maximum number of logically independent values (the total number of observations minus the number of imposed constraints).

Applied to feature selection, chi-square is calculated between every feature variable and the target variable (the occurrence of the feature and the occurrence of the class). If the feature variable is independent of the class, it is discarded; otherwise, it is selected.

### 2.4. Local Binary Patterns

This approach to feature reduction is based on the uniform Local Binary Pattern (LBP), a popular texture descriptor. LBP is defined across each pixel value ($I_c$) on a local circular neighborhood of radius $R$ of size $N$ pixels, thus:

$$LBP(N, R) = \sum_{n=0}^{N-1} s(I_n - I_c) 2^N, \qquad (2)$$

where $s(x) = 1$ if $x \geq 0$; 0 otherwise. A histogram of the resulting binary numbers describes the texture of a given image.

When calculating (2), two types of patterns are distinguished, those with less than three transitions between 0 and 1, known as *uniform* patterns, and the remainder, which are called *nonuniform*.

In this work, $N = 8$ and $R = 1$ and only uniform patterns, as already mentioned, are considered. After LBP extraction from each channel of a CNN layer, dimensionality is reduced with the chi-square feature selection method. In the experimental section, the dimensionality method based on LBP combined with chi-square feature selection is labeled LB.

### 2.5. Deep Co-Occurrence of Deep Representation

A deep co-occurrence representation can be obtained from a deep convolutional layer as proposed in [14]. A co-occurrence is said to occur when the values of two separate activations located inside a given region are greater than a certain threshold. The resulting representation is a tensor with the same dimensions as the activation tensor and can be implemented with convolutional filters.

A convolutional filter can be defined as $F \in \mathbb{R}^{D \times D \times S \times S}$, where $D$ is the number of channels in the activation tensor and where the size of the co-occurrence window is $S = 2 \cdot r + 1$, with $r$ the radius defining the co-occurrence region. Filters are initially set to 1 except for the filter that is related to a given channel; such a filter is initialized to 0 or some very small value $\varepsilon$.

Given the activation tensor $A$ of size $M \times N$ with $D$ channels where $A \in \mathbb{R}^{M \times N \times D}$ and where $A$ is the last convolution operator in a CNN, the co-occurrence tensor $C_T \in \mathbb{R}^{M \times N \times D}$ can be considered as a convolution between the activation tensor after thresholding the co-occurrence filter, thus:

$$C_T = (A_{\rho_A} * F) \cdot \rho_A \qquad (3)$$

where $A_{\rho_A} = A \cdot \rho_A$, with $\rho_A \in \mathbb{R}^{M \times N \times D}$, and $\rho_A = A > \bar{A}$, with $\bar{A}$ the average mean of the activation map produced after the last convolutional layer. In other words, given the activation $\alpha_{i,j}^k$:

$$\rho_A(i, j, k) = \begin{cases} 1, \text{if } \alpha_{i,j}^k > \frac{1}{M \cdot N \cdot D} \sum_{i=1}^{M} \sum_{j=1}^{N} \sum_{k=1}^{D} \alpha_{i,j}^k \\ 0, \text{otherwise.} \end{cases} \qquad (4)$$

For pseudo-code, see [14]. In the experimental section, the representation based on co-occurrence representation is labeled CoOC.

### 2.6. Global pooling measurements

The input to a global pooling layer is a set of $n_{A^{(l)}}$ activation maps computed previously by layer $l$, and the output is one global measurement $g(A_i^{(l)})$ for each activation map

$A_i^{(l)} (1 \leq i \leq n_{A^{(l)}})$. The $A^{(l)}$ measurements then become the inputs to an FC layer. In [18], these pooling measurements are transformed into feature vectors. Two global pooling measurements are used for feature extraction in the experiments presented here: Global Entropy Pooling (GEP) and Global Mean Thresholding Pooling (GMTP).

GEP computes the entropy value of $A_i^{(l)}$. Given the probability distribution $p_i^{(l)}$ of $A_i^{(l)}$, calculated first by normalizing values to [0, 255] and then by computing a histogram from the normalized activation map using 255 bins, $p_i^{(l)}$ is simply the resulting histogram divided by the sum of its elements:

$$\sum_j p_i^{(l)}[j] = 1, (0 \leq j \leq 255) \qquad (5)$$

Thus, GEP is defined as:

$$\text{GEP}(A_i^{(l)}) = -\sum_j p_i^{(l)}[j] \ln(p_i^{(l)}[j]) \qquad (6)$$

Unlike GEP, GMPT includes more layer information into the feature extraction process. To compute GEP, a threshold $T_g^{(l)}$ must be obtained by averaging the value of the entire set of activation maps $A^{(l)}$:

$$T_g^{(l)} = \frac{\sum_i \sum_v \sum_u (A_i^{(l)}[v,u])}{n_{A^{(l)}} * h_{A^{(l)}} * w_{A^{(l)}}} \qquad (7)$$

where $v$ and $u$ are an element's position in the $i$-th activation map computed previously by layer $l$. Whereas $n_{A^{(l)}}$, as already noted, represents the number of activation maps, $h_{A^{(l)}}$ and $w_{A^{(l)}}$ are the height and width of each map. Thus, GMPT is the proportion of elements in each $A^{(l)}$ with values below threshold $T_g^{(l)}$.

*2.7. Sequential Forward Floating Selection of Layers/Classifiers*

In some of the experiments presented in this work, we examine the performance of a layer selection method (i.e., a classifier selection procedure) using Sequential Forward Floating Selection as described in [27]. Selecting classifiers using SFFS is performed by including models in the final ensemble that produce the highest increment of performance compared to an existing subset of models. A backtracking step replaces the worst model from the actual ensemble using the better performing model. Since SFFS requires a training phase to select the best models for the task, we perform a leave-one-out-data set selection protocol.

**3 Experimental Results**

This section describes the experimental results on twelve publicly available medical image data sets:
- CH (CHO data set [28]) contains 327 fluorescence microscope 512×382 images of Chinese Hamster Ovary cells divided into five classes;
- HE (2D HeLa data set [28]) contains 862 fluorescence microscopy 512×382 images of HeLa cells stained with various organelle-specific fluorescent dyes. The images are divided into ten classes of organelles;
- RN (RNAi data set [29]) contains 200 fluorescence microscopy 1024 x1024 TIFF images of fly cells (D. melanogaster) divided into ten classes;
- MA (C. Elegans Muscle Age data set [29]) contains 237, 1600×1200 images for classifying the age of the nematode given twenty-five images of C. elegans muscles collected at four ages;
- TB (Terminal Bulb Aging data set [29]) is the companion data set to MA and contains 970 768×512 images of C. elegans terminal bulbs collected at seven ages;
- LY (Lymphoma data set [29]) contains 375, 1388×1040 images of malignant lymphoma representative of three types;
- LG (Liver Gender Caloric Restriction (CR) data set [29]) contains 265, 1388×1040 images of liver tissue sections from six-month male and female mice on a CR diet;

- LA (Liver Aging Ad-libitum data set [29]) contains 529, 1388×1040 images of liver tissue sections from female mice on an ad-libitum diet divided into four classes representing the age of the mice;
- BGR (Breast Grading Carcinoma [30]): this is a Zenodo data set (record: 834910#.Wp1bQ-jOWUl) that contains 300 1280×960 annotated histological images of twenty-one patients with invasive ductal carcinoma of the breast representing three classes/grades;
- LAR (Laryngeal data set [31]): this is a Zenodo data set (record: 1003200#.WdeQcnBx0nQ) containing 1320, 1280×960 images of thirty-three healthy and early-stage cancerous laryngeal tissues representative of four tissue classes;
- LO (Locate Endogenous data set [32]) contains 502, 768×512 images of endogenous cells divided into ten classes. This data set is archived at https://integbio.jp/dbcatalog/en/record/nbdc00296 ;
- TR (Locate Transfected data set [32]) is a companion data set to LO and contains 553, 768×512 images divided into the same ten classes as LO but with the addition of one more class for a total of eleven classes.

Data sets 1-8 can be found at https://ome.grc.nia.nih.gov/iicbu2008/, data sets 9-10 are on Zenodo and can be accessed by their record number provided in parentheses in the data set descriptions. Data sets 10 and 12 are available upon request.

The five-fold cross-validation protocol is applied to all data sets except for LAR, which uses a three-fold protocol. Although the size of the original images is provided above in the data set descriptions, all images were resized to fit the input size for the given CNN model.

In our experiments, we obtained better results tuning the CNN on each training set without PCA processing and the application of the methods locally (i.e., separately on each channel of a given layer). For this reason, most of the results reported in the following tables for the dimensionality reduction methods (unless otherwise specified) are based on tuning the CNNs without PCA postprocessing and the local application of methods. As noted in the introduction, the Wilcoxon signed-rank test [33] is the measure used to validate experiments.

Reported in Table 1 is the performance of all the approaches using ResNet50, and Reported in Table 2 are the most interesting results using GoogleNet. The row labeled *CNN* in Tables 1 and 2 reports the performance obtained using the standard standalone CNN, be it ResNet50 (Table 1) or GoogleNet (Table 2). The label *TunLayer-x* represents features extracted for SVM training using the $x$ to last layer of the network tuned on the given training set. The best performance for TunLayer-x is obtained with $x = 3$; we also report, for comparison purposes, the performance of Layer-3 on the CNN pre-trained on ImageNet without tuning on the given data sets. The label *TunFusLayer* is the fusion by sum rule of the TunLayer-x classifiers. Row X+Y indicates the sum rule between X and Y, and the method named *g-DC* is DC applied globally as in [17].

The SVM classifiers were tested using LibSVM and fitecoc (MathWorks). Performance was higher using fitecoc (except for CoOC) and when the CNN was not tuned. For the last four layers, we trained SVM using the original features.

Feature vectors produced by a given layer of the CNN with a dimensionality higher than 5000 were processed by applying the dimensionality reduction techniques in the following ways:
- DoOC, GEP, and GMTP use a single value extracted from each channel (see details in the previous section);
- For the other approaches, the method is first applied separately on each channel; then 1000/(Number of Channels )features are extracted from each channel;
- For g-DC, all the features from all channels are first concatenated; then, they are reduced to a 1000-dimension vector by applying DCT.
- In the following Tables, some classifiers are labeled as:

- *Ens15CNN*, the sum rule among fifteen standard ResNet50 CNNs or fifteen standard GoogleNets. This is a baseline approach since our method is an ensemble of classifiers;
- *(DC+GMTP)-2*, the approach *(DC+GMTP)* where the two last layers of the CNN are not used for feeding SVM. Notice that DC and GMTP are extracted considering two different trainings of the CNN (this is done in order to increase the diversity of the features extracted by the two methods);
- SFFS(X) means that we combine by sum rule *X* SVMs.

For GoogleNet, only the most interesting approaches that were reported for ResNEt50 are tested; this is done to reduce computation time.

**Table 1.** Performance using ReNet50.

| Method | ResNet | | | | | | | | | | | | Avg |
|---|---|---|---|---|---|---|---|---|---|---|---|---|---|
| | CH | HE | LO | TR | RN | TB | LY | MA | LG | LA | BG | LAR | |
| CNN | 93.2 | 91.0 | 91.6 | 94.2 | 56.5 | 62.5 | 76.3 | 90.8 | 92.6 | 88.9 | 87.3 | 91.6 | 85.5 |
| Ens15CNN | 99.0 | 96.0 | 98.4 | 98.5 | 66.0 | 70.2 | 92.2 | 95.8 | 99.3 | 98.6 | 94.0 | 94.7 | 91.9 |
| TunLayer-3 | 98.1 | 93.8 | 97.4 | 96.3 | 74.0 | 68.2 | 86.9 | 94.1 | 100 | 98.2 | 91.3 | 91.4 | 91.2 |
| TunLayer-2 | 96.0 | 91.2 | 95.4 | 95.0 | 70.5 | 65.4 | 82.1 | 92.1 | 97.0 | 97.1 | 90.0 | 90.1 | 89.1 |
| TunLayer-1 | 94.1 | 91.7 | 93.4 | 94.9 | 62.0 | 64.9 | 77.0 | 91.6 | 95.3 | 92.5 | 90.0 | 91.2 | 87.3 |
| TunLayer | 94.1 | 91.7 | 93.4 | 94.9 | 62.0 | 64.9 | 77.0 | 91.7 | 95.3 | 92.6 | 90.0 | 91.2 | 87.3 |
| TunFusLayer | 96.6 | 93.0 | 95.4 | 96.0 | 68.5 | 67.0 | 80.8 | 92.9 | 98.3 | 96.7 | 90.6 | 91.1 | 89.5 |
| Layer-3 | 96.3 | 91.7 | 95.4 | 95.3 | 65.0 | 63.8 | 75.7 | 89.2 | 99.0 | 97.7 | 88.3 | 90.4 | 87.9 |
| DC | 99.1 | 95.0 | 98.6 | **97.8** | 83.5 | 73.4 | 90.9 | 97.5 | **100** | 98.9 | 90.7 | 91.6 | 93.1 |
| g-DC | 96.9 | 91.0 | 95.6 | 94.0 | 66.0 | 70.2 | 83.2 | 92.5 | 99.0 | 94.8 | 89.7 | 90.1 | 88.6 |
| PC | 98.1 | 94.8 | 98.4 | 97.3 | 81.5 | 75.3 | 91.5 | 97.5 | 99.7 | 99.4 | 92.3 | **93.4** | 93.3 |
| LB | 98.1 | 93.3 | 98.4 | 96.7 | 78.5 | 70.1 | 90.9 | 93.3 | 99.3 | 98.7 | 92.7 | 93.3 | 91.9 |
| CHI | 97.8 | 93.0 | 94.2 | 92.5 | 67.5 | 74.6 | 82.1 | 90.8 | 98.7 | 96.0 | 92.0 | 91.4 | 89.2 |
| CoOC | 98.1 | 94.6 | 97.2 | 97.5 | 80.5 | 71.1 | 85.9 | 94.2 | 98.7 | 98.8 | 92.3 | 92.4 | 91.8 |
| GMTP | 99.1 | 94.1 | **99.0** | 97.3 | 81.0 | 73.5 | 91.2 | 95.8 | 99.7 | 99.2 | **93.7** | 92.8 | 93.0 |
| GEP | **99.4** | 94.3 | 98.8 | **97.8** | 79.0 | 73.7 | 91.7 | 97.5 | 99.7 | 98.9 | **93.7** | 92.4 | 93.0 |
| DC+PC | 98.8 | 95.2 | 98.6 | 97.6 | 83.5 | 76.6 | 92.5 | **98.8** | 99.7 | **99.6** | 92.3 | 93.0 | **93.9** |
| DC+GMTP | 98.8 | **95.1** | **99.0** | 97.6 | 85.0 | 74.6 | 92.5 | 98.3 | **100** | **99.6** | 92.7 | 92.8 | 93.8 |
| (DC+GMTP)-2 | 99.1 | 94.3 | 98.4 | 98.0 | 87.5 | 74.6 | 92.8 | 97.5 | **100** | 99.4 | 92.3 | 92.6 | 93.9 |
| DC+PC+GMTP | 98.8 | 94.5 | **99.0** | 97.6 | 84.0 | 76.7 | **93.1** | 98.3 | 99.7 | **99.6** | 92.3 | 93.1 | **93.9** |
| (DC+PC+GMTP)-2 | 98.8 | 94.2 | 98.6 | 97.8 | 86.5 | 76.3 | **92.3** | 98.3 | 100 | 99.4 | 92.3 | 93.2 | **94.0** |
| SFFS(20) | 98.8 | 94.5 | 98.8 | 97.5 | 85.0 | **76.8** | **93.1** | 98.3 | 100 | 99.6 | 92.3 | 92.2 | **93.9** |
| SFFS(10) | 98.5 | 94.6 | 98.6 | 97.3 | 84.5 | 73.9 | **93.1** | 99.2 | 100 | 99.4 | 91.7 | 91.1 | 93.5 |

**Table 2.** Performance using GoogleNet.

| Method | GoogleNet | | | | | | | | | | | | Avg |
|---|---|---|---|---|---|---|---|---|---|---|---|---|---|
| | CH | HE | LO | TR | RN | TB | LY | MA | LG | LA | BG | LAR | |
| CNN | 96.3 | 88.4 | 94.4 | 92.7 | 40.5 | 61.2 | 72.0 | 86.7 | 94.3 | 89.5 | 89.3 | 88.3 | 82.8 |
| Ens15CNN | 97.8 | 93.7 | 97.6 | 96.0 | 55.5 | 68.9 | 74.4 | 88.3 | 95.0 | 84.7 | 94.6 | 92.8 | 86.6 |
| TunLayer-3 | 97.8 | 91.4 | 97.2 | 94.4 | 63.0 | 64.3 | 79.7 | 87.1 | 97.7 | 95.4 | 91.0 | 90.7 | 87.5 |
| TunLayer-2 | 97.5 | 90.9 | 96.0 | 93.3 | 64.5 | 64.2 | 77.9 | 84.6 | 97.0 | 95.8 | 92.0 | 91.0 | 87.1 |
| TunLayer-1 | 96.0 | 89.3 | 95.0 | 93.3 | 43.0 | 64.0 | 74.4 | 86.2 | 97.3 | 92.9 | 90.7 | 88.8 | 84.2 |
| TunLayer | 96.0 | 89.3 | 95.0 | 93.3 | 43.0 | 64.0 | 74.4 | 86.2 | 97.3 | 92.9 | 90.7 | 88.9 | 84.2 |
| TunFusLayer | 96.9 | 90.3 | 95.6 | 94.5 | 55.0 | 64.7 | 77.9 | 87.1 | 97.7 | 96.2 | 92.3 | 90.0 | 86.5 |
| DC | 97.5 | 92.3 | 97.2 | 94.0 | 64.0 | 71.2 | 79.7 | 90.8 | 99.7 | 97.7 | 91.7 | 91.7 | 89.0 |
| g-DC | 95.1 | 85.4 | 89.6 | 84.9 | 53.5 | 67.6 | 71.5 | 80.0 | 97.7 | 92.0 | 82.0 | 89.5 | 82.4 |
| PC | 97.2 | 91.9 | 97.0 | 94.5 | 55.5 | 71.5 | 77.6 | 89.2 | 99.0 | 96.9 | 92.7 | 91.8 | 87.9 |
| GMTP | 98.1 | 92.4 | 97.6 | 95.3 | 67.5 | 70.7 | 80.8 | 89.2 | 99.0 | 98.1 | 93.3 | 91.7 | 89.5 |
| GEP | 98.5 | 92.8 | 98.0 | 95.3 | 66.0 | 70.1 | 80.5 | 91.2 | 98.7 | 98.1 | 93.7 | 92.0 | 89.6 |
| DC+PC | 96.9 | 92.4 | 97.6 | 95.3 | 62.0 | 71.9 | 78.9 | 90.0 | 99.7 | 97.3 | 91.7 | 91.6 | 88.8 |
| DC+GMTP | 97.5 | 92.6 | 98.0 | 95.3 | 66.5 | 71.4 | 81.3 | 89.2 | 99.7 | 97.7 | 93.0 | 92.0 | 89.5 |
| (DC+GMTP)-2 | 96.9 | 93.1 | 98.0 | 94.4 | 68.5 | 72.6 | 83.2 | 92.1 | 99.7 | 97.7 | 92.7 | 92.4 | 90.1 |
| DC+PC+GMTP | 96.9 | 92.3 | 98.0 | 95.5 | 64.5 | 71.8 | 79.5 | 89.2 | 99.3 | 97.5 | 93.0 | 91.7 | 89.1 |
| (DC+PC+GMTP)-2 | 97.2 | 93.0 | 97.0 | 94.2 | 63.5 | 72.2 | 81.6 | 90.4 | 99.7 | 97.3 | 92.0 | 92.0 | 89.2 |

The analysis of the results reported in Tables 1 and 2 lead to the following set of observations:
- DC clearly outperforms (p-value 0.0001) g-DC on both GoogleNet and ResNet50. Applying DCT separately on each channel boosts performance with respect to a single application of DCT on the whole layer.
- The best methods for reducing the dimensionality of the inner layers are DC, PC, GMTP, and GEP.
- On average, the best approach is given by (DC+GMTP)-2, i.e., by the sum rule between DC and GMTP.
- On average, discarding the SVMs trained with the two last layers slightly improves performance.
- DC outperforms (p-value 0.01) on any TunLayer-x; this implies that the inner layers are also useful on the tuned networks.

Both TunLayer-3 and DC strongly outperform (p-value 0.01) CNN on all the tested data sets. Using a GoogleNet/ResNet50 directly to classify images does not maximize performance, probably due to overfitting given the size of the training sets. Notice that we have trained an SVM classifier on each of the ten layers. Considering the size of GoogleNet and ResNet50, using larger CNNs with so many layers would not be the best choice.

To test the generalizability of our approach, Table 3 reports experiments run on a popular Virus benchmark data set [34] located at http://www.cb.uu.se/_Gustaf/virustexture/. This data set contains 1500, 41×41 Transmission Electron Microscopy (TEM) images of viruses belonging to fifteen species of viruses and is divided into two different data sets: 1) the *object scale* data set, so named because the radius of every virus in each image is 20 pixels, and 2) the *fixed scale* data set, so called because each virus image is represented such that the size of 1 pixel corresponds to 1nm. The first data set, used in the following experiments, is publicly available. The second is proprietary, so it is unavailable for testing due to copyright issues. It is the object scale data set that is widely reported in the literature.

Regarding object scale, two networks were trained: DenseNet201, that is the network providing the best performance on the object scale dataset in the literature, and ResNet50, because a large number of relevant papers report results using this network. Both CNNs were trained for fifty epochs, with all other parameters the same as those noted in the tests reported above.

**Table 3.** Performance in the Virus data set to test generalizability.

| Method | | DenseNet201 | ResNet50 | DenseNet201+ResNet50 |
|---|---|---|---|---|
| CNN | --- | 81.60 | 77.13 | 82.53 |
| TunLayer-3 | LibSVM | 86.73 | 81.47 | 86.00 |
| TunLayer-2 | | 84.07 | 81.00 | 84.67 |
| TunLayer-1 | | 81.67 | 79.80 | 83.33 |
| TunLayer | | 81.67 | 79.80 | 83.33 |
| TunLayer-3 | FitEcoc | 85.67 | 83.73 | 85.67 |
| TunLayer-2 | | 83.00 | 80.80 | 84.07 |
| TunLayer-1 | | 81.00 | 79.27 | 81.67 |
| TunLayer | | 81.00 | 79.27 | 81.67 |
| DC-2 | | 87.20 | 87.73 | 89.07 |
| DC-2 | FitEcoc | 86.73 | 86.73 | 88.13 |
| (DC+GMTP)-2 | LibSVM | 89.27 | 88.27 | **89.60** |
| (DC+GMTP)-2 | FitEcoc | 87.67 | 88.73 | 88.67 |
| (DC+PC+GMTP)-2 | LibSVM | 88.93 | --- | --- |
| (DC+PC+GMTP)-2 | FitEcoc | 87.67 | --- | --- |

In Table 3, we report the performance obtained using both LibSVM and fitecoc classifiers. DenseNet produces a better performance using LibSVM. To reduce computation time, the combination (DC+PC+GMTP)-2 is not run on ResNet because, when coupled with DenseNet, it obtains a performance similar to (DC+GMTP)-2. The last column of Table 3 reports the fusion by sum rule between the two CNNs before the sum of the scores of each ensemble is normalized by dividing the score by the number of trained SVMs.

Finally, in Table 4, we compare our approach with the best performance reported in the literature. As can be observed in Table 4, our proposed method obtains state-of-the-art performance. In [34], the reported performance is obtained using the *fixed scale* data set. Since that data set is not publicly available, comparisons with [34] cannot be made. By combining features computed on the *object scale* with the *fixed scale* data sets, an accuracy of 87.0 % is obtained in their work.

**Table 4.** Comparison with the literature

| This work | [35] | [36] | [37] | [38] | [34] | [39] | [40] |
|---|---|---|---|---|---|---|---|
| 89.60 | 89.47 | 89.00 | 88.00 | 87.27 | 87.00* | 86.2 | 85.7 |

Note: the method notated with * combines descriptors based on both *object scale* and *fixed scale* images.

**5. Conclusion**

The objective of this work was to explore the power of using both the intermediate and the last layers of three pre-trained CNNs to evaluate features with fixed-length feature vectors that can be used to train an ensemble of SVMs. To overcome the high dimensionality of the features extracted from inner layers, experiments tested many different dimensionality reduction techniques, including two classic feature transforms (DCT and PCA), a feature selection approach (chi-square feature selection), a texture descriptor (local binary pattern) followed by feature selection and a representation based on the co-occurrence among elements of the channels of inner layers.

The best ensemble reported here is shown to significantly boost the performance of standard CNN on a large and diverse group of image data sets as well as on a popular benchmark virus data set where the ensemble obtained state-of-the-art performance.

As future works, we plan on combining this approach with other deep neural networks and to test different methods for representing the inner layers in a compact way for feeding SVM.

**Author Contributions:** L.N. conceived the presented idea., L.N. performed the experiments. S.B., L.N. and S.G. wrote the manuscript and S.B. provided some resources.

**Acknowledgments:** We acknowledge the support of NVIDIA's GPU Grant Program and the donation of a TitanX GPU for training the deep learner in this work.

**Conflicts of Interest:** The authors declare no conflict of interest.